\documentclass[runningheads]{llncs}

\usepackage{accv}

\usepackage{accvabbrv}

\usepackage[accsupp]{axessibility}  
\usepackage[breaklinks,colorlinks,citecolor=accvblue]{hyperref}

\usepackage{xcolor}
\usepackage{multirow}%
\usepackage{xcolor}%
\usepackage{textcomp}%
\usepackage{manyfoot}%
\usepackage{booktabs}
\usepackage{listings}%
\usepackage{multirow}
\usepackage{booktabs}
\usepackage{pifont}
\usepackage{soul}
\usepackage{enumitem}
\usepackage{tabularx}
\usepackage{booktabs}
\usepackage{lipsum}
\usepackage{tabularray}
\usepackage{mathbbol}
\usepackage[utf8]{inputenc}
\usepackage{color, colortbl}
\usepackage{booktabs}
\usepackage{graphicx}

\newcommand{\xmark}{\ding{55}}%

\begin{document}

\title{Amodal Instance Segmentation with Diffusion Shape Prior Estimation} 

\titlerunning{Abbreviated paper title}

\author{Minh Tran\inst{1} \and
Khoa Vo\inst{1} \and
Tri Nguyen\inst{2} \and
Ngan Le\inst{1}}

\authorrunning{M. Tran et al.}
\titlerunning{AISDiff}

\institute{University of Arkansas, Fayetteville AR, USA \and
Coupang, Inc., Seattle WA, USA
\\
{\small \url{https://uark-aicv.github.io/AISDiff}}
}

\maketitle

\begin{abstract}
Amodal Instance Segmentation (AIS) presents an intriguing challenge, including the segmentation prediction of both visible and occluded parts of objects within images. Previous methods have often relied on shape prior information gleaned from training data to enhance amodal segmentation. However, these approaches are susceptible to overfitting and  disregard object category details. Recent advancements highlight the potential of conditioned diffusion models, pretrained on extensive datasets, to generate images from latent space. Drawing inspiration from this, we propose AISDiff with a Diffusion Shape Prior Estimation (DiffSP) module. AISDiff begins with the prediction of the visible segmentation mask and object category, alongside occlusion-aware processing through the prediction of occluding masks. Subsequently, these elements are inputted into our DiffSP module to infer the shape prior of the object. DiffSP utilizes conditioned diffusion models pretrained on extensive datasets to extract rich visual features for shape prior estimation. Additionally, we introduce the Shape Prior Amodal Predictor, which utilizes attention-based feature maps from the shape prior to refine amodal segmentation. Experiments across various AIS benchmarks demonstrate the effectiveness of our AISDiff.
\end{abstract}    
\section{Introduction}
Amodal perception, as described in \cite{kellman1991theory}, describe human's remarkable ability to perceive objects in their entirety despite occlusion. Building upon this concept, the pioneering studies by ~\cite{zhu2017semantic,li2016amodal} introduced amodal instance segmentation (AIS). This approach aims to predict the complete shape of objects, encompassing both their visible and occluded regions. Indeed, AIS exhibits vast potential across various domains, as evidenced by its applications in robot manipulation \cite{back2022unseen} and autonomous driving \cite{qi2019amodal}.
Across various AIS benchmarks \cite{follmann2019learning, qi2019amodal, zhu2017semantic}, a multitude of approaches addressing the AIS challenge have emerged in the literature. These approaches, as evidenced by numerous studies \cite{li2016amodal, follmann2019learning, qi2019amodal, mohan2022amodal, tran2022aisformer, jang2020learning, tran2024shapeformer,xiao2020amodal}, demonstrate the ongoing efforts to tackle this challenge.

\begin{figure}[!t]
    \centering
    \includegraphics[width=.85\textwidth]{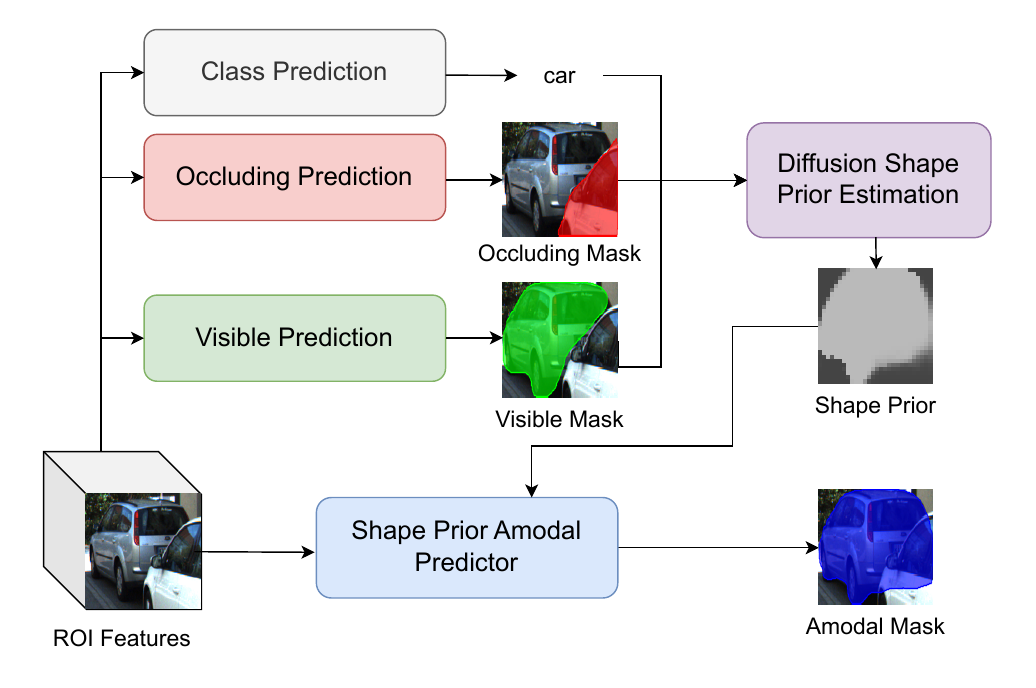}
    \caption{Overall architecture of AISDiff. AISDiff predicts the visible segmentation mask and the object category while simultaneously addressing occlusion by predicting the occluding mask. Next, these predictions are integrated into the Diffusion Shape Prior Estimation (DiffSP) module to establish the object's shape prior. This shape prior is then utilized by AISDiff to produce the amodal segmentation.} 
    \label{fig:teaser}
\end{figure}

Recent research \cite{duncan1984selective, yao2022self, xiao2021amodal, jang2020learning, gao2023coarse} highlights the effectiveness of integrating shape prior information in AIS. Indeed, These shape prior AIS methods typically construct shape-prior knowledge from the training dataset, which is later utilized to train the AIS model. In \cite{xiao2021amodal}, for instance, the authors employ variational autoencoders to reconstruct amodal masks. The concept revolves around using ground truth amodal masks, utilizing autoencoders to reconstruct them, and storing the encoded codebooks as shape priors. Similarly, in \cite{gao2023coarse}, the authors also construct a shape prior codebook but employ a vector-quantization variational autoencoder. After establishing the shape prior, these method first predict the coarse amodal segmentation and refine the final amodal segmentation mask using the built shape prior. 
However, there are limitations to these approaches. Firstly, the shape prior tends to overfit to the training data, consequently leading to overfitting in amodal mask prediction overall. Secondly, since the shape prior is built solely from ground truth amodal masks, it may overlook the object category, which could provide significant supplementary information for deriving the shape prior.

To tackle these issues, we desgin a \textbf{AIS} mask head with \textbf{Diff}usion Shape Prior Estimation (\textbf{AISDiff}). The design of AISDiff is depicted in Figure \ref{fig:teaser}. In essence, AISDiff begins by predicting the visible segmentation mask and the category of the object of interest. Simultaneously, it conducts occlusion-aware processing by predicting the occluding mask, which is the segmentation of occluding elements within the specified ROI. Subsequently, these three pieces of information are fed into the proposed Diffusion Shape Prior Estimation (DiffSP) module to derive the shape prior of the object. Finally, leveraging this shape prior, AISDiff generates the amodal segmentation. 

Specifically, DiffSP leverages the successes of conditioned diffusion models (such as Stable Diffusion \cite{rombach2022high} and GLIDE \cite{nichol2021glide}), which are pretrained on extensive language vision datasets like LAION \cite{schuhmann2021laion}. This enables the model to capture rich visual features, making it suitable as prior knowledge for downstream tasks \cite{zhan2023amodal, ozguroglu2024pix2gestalt}.
Building upon this foundation, we feed a trained conditioned diffusion model with an ROI image containing only the visible pixels of the object of interest, expecting the model to generate the missing parts. Additionally, an occluding mask and a textual description of the object category is also feed to condition the mdoel. Subsequently, the denoising process iterates $T$ steps to output the generated image containing the occluded parts.
However, rather than relying on the final generated pixels, DiffSP exploits on the attention mechanism between the conditioning information and the image features. This attention map remains relatively stable across time steps, thereby reducing the denoising time needed to obtain the shape prior.
Furthermore, we design the Shape Prior Amodal Predictor, which learns the attention-based amodal feature map from the acquired shape prior to predict the amodal mask segmentation.

In summary, our contributions are as follows:
\begin{itemize}
\item We present AISDiff, a novel AIS mask head featuring a Diffusion Shape Prior Estimation module. This model predicts the visible segmentation mask and category of the object while considering occlusion. It then uses these predictions to estimate the shape prior of the object before generating the final amodal segmentation mask.

\item We propose DiffSP module, harnessing the efficacy of conditioned diffusion models to derive the shape prior of the object of interest.

\item We introduce the Shape Prior Amodal Predictor, which learns attention-based amodal feature maps from the obtained shape prior to predict the amodal segmentation.
\end{itemize}
\section{Related Work}
\noindent
\subsection{Amodal Segmentation}
Amodal instance segmentation involves predicting an object’s
shape, including both its visible and occluded parts.
Li and Malik \cite{li2016amodal} pioneered a method aimed at addressing AIS. They proposed enlarging the modal bounding box in alignment with high heatmap values and synthesizing occlusions. Following this seminal work, various methodologies have surfaced in literature. Notably, ORCNN \cite{follmann2019learning} introduces instance mask heads for both amodal and visible instances, along with an additional head for predicting occluded masks. ASN \cite{qi2019amodal} builds upon ORCNN by integrating a multi-level coding module for bidirectional feature modeling of visible and amodal aspects. BCNet \cite{ke2021deep} enhances amodal mask prediction by incorporating a supplementary branch dedicated to predicting occlusion masks within the bounding box. AISFormer \cite{tran2022aisformer} introduces a transformer-based mask head, demonstrating the efficacy of transformer modeling in generating AIS masks. However, their approach, which consolidates all mask relationships into one transformer model, leads to compromised visible segmentation output, consequently affecting the quality of amodal segmentation output due to bidirectional feature relations as mentioned earlier.

Recent studies \cite{xiao2021amodal,jang2020learning,gao2023coarse} underscore the benefits of integrating shape priors into AIS. These methods leverage prior knowledge of mask shapes to improve amodal mask predictions. VRSP-Net \cite{xiao2021amodal} predicts coarse amodal masks, retrieves shape priors using a simple autoencoder, and then refines the final amodal mask predictions. AmodalBlastomere \cite{jang2020learning} employs a similar strategy with a variational autoencoder for blastomere and cell segmentation. C2F-Seg\cite{gao2023coarse} constructs a shape prior codebook using a vector-quantization variational autoencoder. After establishing the shape prior, the method first predicts a coarse amodal segmentation. This coarse segmentation is then refined to produce the final amodal segmentation mask using the built shape prior. Despite their progress, these methods often overlook the importance of object categories when utilizing prior shapes. Moreover, their training procedures frequently lead to overfitting of the shape prior model to the training dataset. Additionally, these approaches simply incorporate the shape prior by concatenating it with visible features to refine amodal masks.

\subsection{Diffusion Models} \label{related:diffusion}
The Denoising Diffusion Probabilistic Model (DDPM)\cite{ho2020denoising} has become a widely used generative architecture in computer vision. Its popularity stems from its ability to model multi-modal distributions, training stability, and scalability. The study by \cite{dhariwal2021diffusion} first showed that diffusion models outperform GANs\cite{goodfellow2014generative} in image synthesis. To enhance computational efficiency, Stable Diffusion~\cite{rombach2022high}, trained on LAION-5B~\cite{schuhmann2022laion}, applied a diffusion model in the latent space of a variational autoencoder~\cite{kingma2013auto}. Subsequently, major improvements were made to boost diffusion model performance~\cite{ho2022classifier,song2020denoising}. With the release of Stable Diffusion~\cite{rombach2022high} as a powerful generative tool, many works have adapted it to tackle tasks in various domains such as image editing~\cite{brooks2023instructpix2pix,gal2022image,ruiz2023dreambooth} and image segmentation~\cite{xu2023open,amit2021segdiff,baranchuk2021label}. Recently, diffusion models have been applied to the problem of amodal completion. Notably, \cite{ozguroglu2024pix2gestalt} and \cite{zhan2024amodal} leverage diffusion models, such as Stable Diffusion \cite{rombach2022high}, to train on proposed amodal completion datasets with synthetic occlusion. Additionally, \cite{xu2024amodal} utilize pretrained features from Stable Diffusion \cite{rombach2022high} for their UNet model aimed at amodal mask completion. Unlike these works, AISDiff is an AIS framework designed to amodally detect and segment instances in images. Furthermore, AISDiff takes advantage of the attention maps in diffusion models to build prior knowledge without denoising to the final output.

\section{Method}
\subsection{Overall AIS Setup}
\label{sec:ais_setup}
Given an input image $\mathbf{I}$, we follow most of previous AIS settings \cite{xiao2021amodal, follmann2019learning, ke2021deep, tran2022aisformer}, utilizing a pre-trained backbone network, such as ResNet \cite{he2017mask}, RegNet \cite{schneider2017regnet} to extract spatial visual representation. An object detector such as FCOS \cite{tian2019fcos}, or Faster-RCNN\cite{he2017mask},
can be subsequently adopted to obtain $n$ regions of interest (RoI) predictions and their corresponding visual features $\{\mathbf{F}^i\}_{i=1}^n$. Follow most of previous works \cite{xiao2021amodal, ke2021deep,tran2022aisformer}, the object detector being chose is Faster R-CNN for fair comparison. Here, each RoI is presented by its visual feature $\mathbf{F}^i \in \mathbb{R}^{C_e\times H_r\times W_r}$, where $C_e$ denotes the feature channel size and $H_r \times W_r$ represents the spatial shape of the pooling feature. 
In this context, given a RoI, AISDiff takes  $\mathbf{F}^i$ as input and aims to predict the amodal mask $\mathbf{M}_a^i$.
Moreover, in this case, we also denote the visible mask $\mathbf{M}_v^i$, and the occluding mask $\mathbf{M}_o^i$.

\begin{figure}[!t]
    \centering
    \includegraphics[width=\columnwidth]{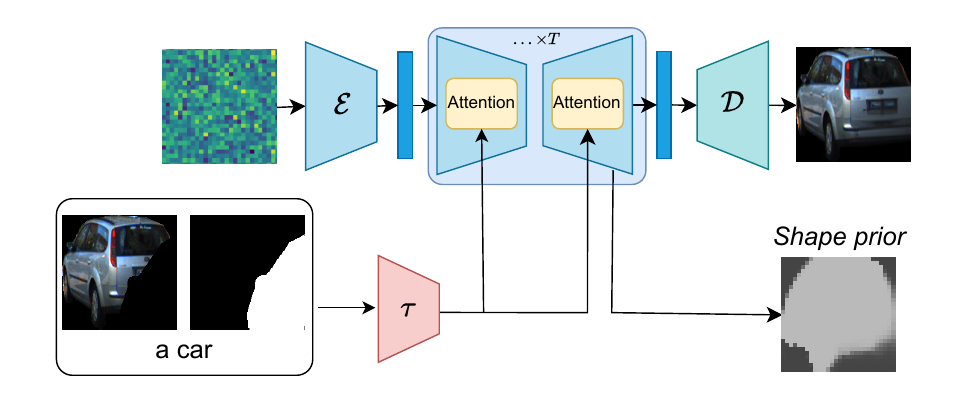}
    \caption{Overall process of Diffusion Shape Prior Estimation (DiffSP). } 
    \label{fig:diffusion_shapeprior}
\end{figure}

\subsection{AISDiff} 
The overall design of AISDiff is depicted in \cref{fig:teaser}. Initially, we discuss the prediction process for the visible segmentation of the object of interest, along with its categories, incorporating occlusion-awareness through the prediction of occluding masks (\cref{sec:vis_occ}). Following this, we introduce the DiffSP method in detail (\cref{sec:DiffSP}). Lastly, we present the Shape Prior Amodal Predictor (\cref{sec:sp_amodal}).

\subsubsection{Occlusion-aware Visible Segmentation}
\label{sec:vis_occ}
Given the ROI feature $\mathbf{F}^i$, AISDiff first aims to predict the visible segmentation mask and the category of the object of interest, while simultaneously conducts occlusion-aware ability by predicting the occluding mask, which is the segmentation of occluding elements within the specified ROI. 
BCNet~\cite{ke2021deep} is utilized as the foundation for the Occlusion-aware Visible Segmentation module. This module consists of two branches: one for occluding mask prediction and the other for visible mask prediction. Drawing from the methodology outlined in ~\cite{ke2021deep}, both branches follow a similar design structure, encompassing two main components: feature extraction and mask prediction.
The feature extraction segment comprises a sequence of layers, including a $3\times3$ convolutional layer with a stride of 1, a Graph Convolutional Network (GCN)~\cite{kipf2016semi} block, and another $3\times3$ convolutional layer with a stride of 1. Subsequently, the mask prediction component is constructed with a $2 \times 2$ transposed convolutional layer employing a stride of 2, coupled with a $1 \times 1$ convolutional layer using a stride of 1.

Furthermore, to enhance occlusion awareness and subsequently improve visible segmentation accuracy, features extracted from the occluding branch are incorporated into the ROI feature $\mathbf{F}^i$ before being fed into the feature extraction section of the visible branch. Simultaneously, features extracted from the visible branch are utilized for object category prediction. This classification step employs a fully connected layer with an output dimension corresponding to the number of categories present in the datasets under consideration.
In summary, the final output of this module comprises the visible mask $\mathbf{M}_a^i$, the occluding mask $\mathbf{M}_o^i$, and the object category $c^i$.

\subsubsection{DiffSP}
\label{sec:DiffSP}
The process depicted in \cref{fig:diffusion_shapeprior} illustrates the Shape Prior Estimation (DiffSP) module. DiffSP builds upon the successes of conditioned diffusion models, such as Stable Diffusion \cite{rombach2022high} and GLIDE \cite{nichol2021glide}, which are pre-trained on comprehensive language-vision datasets like LAION \cite{schuhmann2021laion}. This pre-training equips the model with the ability to capture intricate visual features, rendering it suitable as prior knowledge for subsequent tasks \cite{zhan2023amodal, ozguroglu2024pix2gestalt}.
Expanding on this foundation, DiffSP utilizes a trained conditioned diffusion model and inputs a ROI image containing only the visible pixels of the object, an occluding mask and a textual description of the object category under consideration, expecting the model to generate the obscured parts. Subsequently, the denoising process iterates $T$ steps to produce the generated image containing the occluded regions.
However, instead of relying solely on the final generated pixels, DiffSP capitalizes on the attention mechanism between the conditioning information and the image features. 

Specifically, Stable Diffusion \cite{rombach2022high} is employed as the pre-trained conditioned diffusion model, leveraging its self and cross-attention layers.
Specifically, the random Gausian noise is encoded into latent space and then experiences the denoising process over $T$ time steps to generate the inpainting image. In fact, the ROI image containing only the visible pixels of the object of interest, the occluding mask, and the textual description of the object category serve as conditions and are represented as $y$, which is projected by $\tau$ into an intermediate representation $\tau(y)$.
At each denoising step $t$, a UNet architecture with $L$ layers of self and cross-attention transforms $z_{t}$ into $z_{t-1}$. Specifically, at layer $l$ and time step $t$, the cross-attention layer captures the relationship between $z_t$ and the encoded condition $\tau(y)$, reflecting the entire reconstructed shape of the object. This relationship is formalized as follows: at layer $l$ and time step $t$, the self-attention map is denoted as $\mathcal{A}_S^{l, t}$, and the cross-attention map is denoted as $\mathcal{A}_C^{l, t}$.
Moreover, as demonstrated in \cite{nguyen2024dataset}, the attention map remains relatively stable across time steps. Following the methodology of \cite{nguyen2024dataset}, we average these cross and self-attention maps over layers and time steps, setting $T = 10$.
Additionally, as also suggested in \cite{nguyen2024dataset}, although the cross-attention maps $\mathcal{A}_{C}$ already outline the shape of the reconstructed object, they tend to be coarse-grained and noisy. To refine the precision of object localization, we follow \cite{nguyen2024dataset}, utilizing the self-attention map $\mathcal{A}_{S}$ to enhance $\mathcal{A}_{C}$. Consequently, the shape prior is obtained by: $\mathbf{M}_{sp}= (\mathcal{A}_S)^{\tau} \cdot \mathcal{A}_C$.

\begin{figure}[!t]
    \centering
    \includegraphics[width=.8\textwidth]{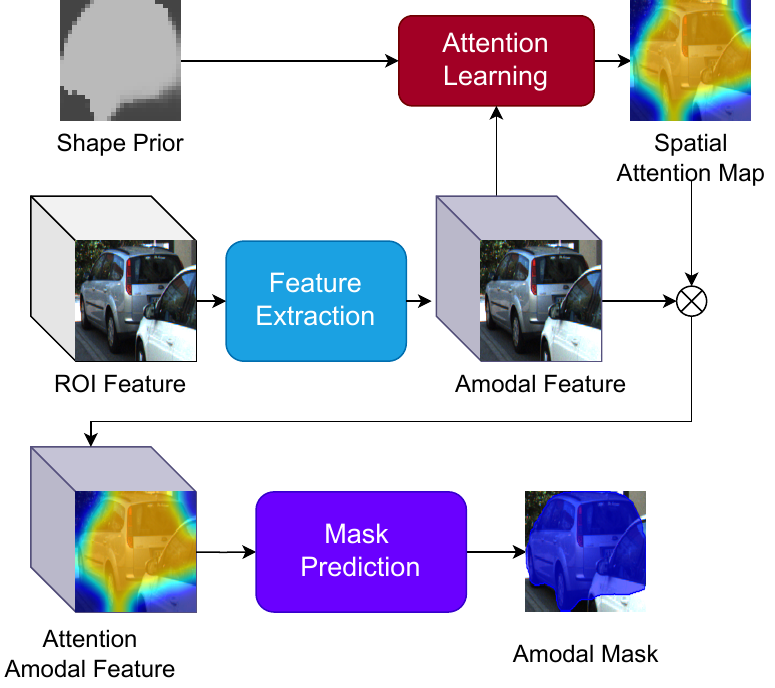}
    \caption{Overall design of Shape Prior Amodal Predictor.} 
    \label{fig:sp_amodal}
\end{figure}

\subsubsection{Shape Prior Amodal Predictor}
\label{sec:sp_amodal}
The design of Shape Prior Amodal Predictor is depicted in \cref{fig:sp_amodal}. Initially, the feature extraction module utilizes the ROI feature $\mathbf{F}^i$ to generate the amodal feature. This module is constructed using a sequence of $3\times 3$ convolutional layers with a stride of 1.
Subsequently, the obtained amodal feature undergoes processing in the attention learning module in conjunction with the shape prior $\mathbf{M}_{sp}$ obtained from DiffSP, aimed at learning the spatial attention map. Specifically, the attention computation involves passing the amodal feature through a sequence of $3\times 3$ convolutional layers with a stride of 1, followed by a sigmoid activation function. This computed attention map is then multiplied with the shape prior $\mathbf{M}_{sp}$.
The spatial attention map is further multiplied with the amodal feature to obtain the attention amodal feature. This feature is then fed into a mask prediction module, which is structured with a $2 \times 2$ transposed convolutional layer employing a stride of 2, coupled with a $1 \times 1$ convolutional layer using a stride of 1, to derive the amodal mask $\mathbf{M}^i_a$

\subsection{Objective Function \& Training}

Employing AIS protocols, the training adopts a two-stage instance segmentation process similar to Mask R-CNN, facilitating concurrent training of both bounding box and amodal mask prediction heads alongside the object detection framework. In essence, the training procedure optimizes a multi-task loss function $\mathcal{L}$ as follows:
\begin{align}
    \mathcal{L} = \mathcal{L}_{det} + \mathcal{L}_{{cls}} + \mathcal{L}_v +\mathcal{L}_o + \mathcal{L}_a
\end{align}
where $\mathcal{L}_{det}$ is object detection loss, defined similarly to that in Faster R-CNN object detection. The occluding mask loss $\mathcal{L}_o$, the visible mask loss $\mathcal{L}_v$, the amodal mask loss $\mathcal{L}_a$, and the classification loss $\mathcal{L}_{cls}$ are computed using cross entropy loss with the corresponding ground truth.

\section{Experiments}
\subsection{Datasets, Metrics and Implementation Details}
\noindent
\textbf{Datasets:}
We benchmark our AISDiff on three AIS datasets, namely KINS \cite{qi2019amodal}, COCOA-cls \cite{follmann2019learning}, and D2SA \cite{follmann2019learning}. KINS is a large-scale traffic dataset with 95,311 training instances and 92,492 testing instances with 7 categories. COCOA-cls is an AIS dataset that is derived from MSCOCO \cite{lin2014microsoft} with 80 categories of 6,763 training instances and 3,799 testing instances. D2SA is an AIS dataset with 60 categories of instances related to supermarket items with 13,066 training instances and 15,654 testing instances.

\noindent
\textbf{Metrics:} 
Following existing AIS methods \cite{xiao2021amodal, tran2022aisformer}, we adopt mean average precision (AP) and mean average recall (AR).

\noindent
\textbf{Implementation details: } We develop AISDiff utilizing the Detectron2 framework \cite{wu2019detectron2}. For the KINS dataset, we employ an SGD optimizer \cite{ruder2016overview} with a learning rate of $0.0025$ and a batch size of 1, over $48,000$ iterations. For the D2SA dataset, training is conducted with an SGD optimizer, a learning rate of $0.005$, and a batch size of 2 over $70,000$ iterations, while for the COCOA-cls dataset, training involves $10,000$ iterations with a learning rate of $0.0005$ and a batch size of 2. All experiments are performed using a Quadro RTX 8000 GPU.

\subsection{Baselines}
We compare AISDiff with state-of-the-art (SOTA) AIS methods, including ORCNN\cite{follmann2019learning}, BCNet\cite{ke2021deep}, VRSP-Net\cite{xiao2021amodal}, and C2F-Seg\cite{gao2023coarse}. These methods share the same AIS setup described in Section \ref{sec:ais_setup}. It is important to note that recent works such as \cite{ozguroglu2024pix2gestalt}, \cite{xu2024amodal}, and \cite{zhan2024amodal} focus on amodal completion with the object of interest already given, and then extract the amodal mask based on this completion. This differs from the AIS framework mentioned, which involves amodal detection and segmentation of instances within images. Therefore, we exclude these methods from our comparison.

\begin{table}[!ht]
    \caption{\textit{Performance comparison on KINS test set} with various backbones. $\dagger$ indicates our reproduced results.}
    \centering
    \setlength{\tabcolsep}{2pt}
    \renewcommand{\arraystretch}{.9}
    \resizebox{.9\textwidth}{!}{
    \begin{tabular}{c|l|c|c|cccc}
    \toprule
         \multicolumn{2}{l|}{\multirow{2}{*}{\textbf{Backbones\& Methods}}}  & \multirow{2}{*}{\textbf{Venue}}&\textbf{{Shape}} & \multirow{2}{*}{$AP\uparrow$} & \multirow{2}{*}{$AP_{50}\uparrow$} & \multirow{2}{*}{$AP_{75}\uparrow$} & \multirow{2}{*}{$AR\uparrow$} \\ 
        ~ & ~ & ~ & \textbf{{Prior}}&  &  &  &  \\ \midrule
        \multirow{7}{*}{\rotatebox{90}{ResNet-50}} & Mask R-CNN\cite{ke2022mask} & ICCV17 & \xmark  & $30.0$ & $54.5$ & $30.1$ & $19.4$ \\  
        ~ & ORCNN\cite{follmann2019learning} & WACV19  &  \xmark & $30.6$ & $54.2$ & $31.3$ & $19.7$ \\ 
        ~ & ASN\cite{qi2019amodal} & CVPR19  & \xmark  & $32.2$ & - & - & - \\ 
        ~ & AISFormer\cite{tran2022aisformer} & BMVC22  & \xmark  & ${33.8}$ & ${57.8}$ & ${35.3}$ & ${21.1}$ \\ 
        ~ & AmodalBlastomere\cite{jang2020learning} & TMI20   & \checkmark & $30.3$ & - & - & - \\ 
        ~ & VRSP-Net\cite{xiao2021amodal}& AAAI21  & \checkmark  & $32.1$ & $55.4$ & $33.3$ & $20.9$ \\ 
        ~ & C2F-Seg \cite{gao2023coarse}& ICCV23  & \checkmark  & $\textbf{36.5}$ & \underline{58.2} & \underline{37.0} & \textbf{22.1} \\ \cmidrule{2-8}
        ~ & \textbf{AISDiff (Ours)}& -  &  \checkmark  & \underline{36.3} & \textbf{58.8} & \textbf{37.2} & \underline{22.0} \\  \midrule
        \multirow{5}{*}{\rotatebox{90}{ResNet-101}} & Mask R-CNN\cite{he2017mask} $\dagger$& ICCV17  & \xmark  & $30.2$ & $54.3$ & $30.4$ & $19.5$ \\ 
        ~ & BCNet\cite{ke2021deep} & CVPR21  & \xmark & $28.9$ & - & - & - \\ 
        ~ & BCNet\cite{ke2021deep}  $\dagger$ & CVPR21 &  \xmark  & $32.6$ & $57.2$ & $35.4$ & $21.5$ \\ 
        ~ & AISFormer\cite{tran2022aisformer}& BMVC22  & \xmark  & ${34.6}$ & ${58.2}$ & ${36.7}$ & ${21.9}$ \\
        ~ & C2F-Seg\cite{gao2023coarse}$\dagger$& ICCV23  & \checkmark  & $\textbf{36.9}$ & $\underline{58.9}$ & $\textbf{37.8}$ & $\textbf{23.1}$ \\ \cmidrule{2-8}
        ~ & \textbf{AISDiff (Ours)}& -  &  \checkmark  & \textbf{36.9} & \textbf{59.6} & \underline{37.5} & \underline{23.0} \\ 
        \bottomrule
    \end{tabular}
    }
    \label{tab:kins}
\end{table}

\subsection{Performance Comparison}
\subsubsection{Quantitative Results} 
\noindent
\textbf{KINS.} \cref{tab:kins} depicts the comparison between AISDiff and SOTA AIS methods on the KINS dataset. AISDiff demonstrates consistent improvements across various backbones, including ResNet-50~\cite{he2016deep} and ResNet-101~\cite{he2016deep}. Specifically, when compared to methods utilizing ResNet-50 as the backbone, AISDiff achieves comparable results with the SOTA method (i.e., and C2F-Seg~\cite{gao2023coarse} 
Simlarly when ResNet-101 is utilized as the backbone, our method achieves great performance, compatible with C2F-Seg.\\
\noindent
\textbf{D2SA.} \cref{tab:d2sa} further validates our approach on D2SA dataset. We achieve best results across all metrics. Specifically, we gains $0.13$ on AP and 0.1 AR in comparison with the SOTA method, i.e. C2F-Seg\cite{gao2023coarse}.\\
\noindent
\textbf{COCOA-cls.} \cref{tab:cocoa-cls} shows our results on COCOA-cls dataset.
AISDiff also outperform other methods on all metrics. In fact, it outperforms the second best by $0.16$ AP and $0.03$ AR.

\begin{table}[!t]
    \caption{\textit{Performance comparison on D2SA test set} with ResNet-50 as backbone. $\dagger$ indicates our reproduced results.}
    \vspace{-0.5em}
    \centering
    \setlength{\tabcolsep}{4pt}
    \renewcommand{\arraystretch}{.9}
    \resizebox{0.9\textwidth}{!}{
    \begin{tabular}{l|c|c|cccc}
    \toprule
         \multirow{2}{*}{\textbf{Methods}} & \multirow{2}{*}{\textbf{Venue}} & \multirow{2}{*}{\textbf{\shortstack{Shape \\ Prior}}} &  \multirow{2}{*}{$AP\uparrow$} & \multirow{2}{*}{$AP_{50}\uparrow$} & \multirow{2}{*}{$AP_{75}\uparrow$} & \multirow{2}{*}{$AR\uparrow$} \\
  ~ & ~ & ~  &  &  &  &  \\ \midrule
                Mask R-CNN\cite{he2017mask}  & ICCV17 & \xmark & $63.57$ & $83.85$ & $68.02$ & $65.18$ \\ 
        ORCNN\cite{follmann2019learning}  & WACV19 & \xmark  & $64.22$ & $83.55$ & $69.12$ & $65.25$ \\ 
        ASN\cite{qi2019amodal}  $\dagger$ & CVPR19 & \xmark & $63.94$ & $84.35$ & $69.57$ & $65.20$ \\ 
        BCNet\cite{ke2021deep}  $\dagger$ & CVPR21 & \xmark & $65.97$ & $84.23$ & $72.74$ & $66.90$ \\ 
        AISFormer\cite{tran2022aisformer}  & BMVC22 & \xmark  & $67.22$ & $84.05$ & $72.87$ & $68.13$ \\ 
        VRSP-Net\cite{xiao2021amodal} & AAAI21 & \checkmark  & ${70.27}$ & $\underline{85.11}$ & ${75.81}$ & ${69.17}$ \\ 
        C2F-Seg\cite{xiao2021amodal}$\dagger$ & ICCV23 & \checkmark  & $\underline{70.88}$ & ${85.07}$ & $\underline{75.85}$ & $\underline{69.19}$ \\ \midrule
        \textbf{AISDiff (Ours)} & - & \checkmark &  $\textbf{71.01}$ & $\textbf{85.12}$ & $\textbf{76.23}$ & $\textbf{69.29}$ \\ \bottomrule
    \end{tabular}
    \vspace{-0.5em}
    }
    \label{tab:d2sa}
\end{table}

\begin{table}[!t]
    \caption{\textit{Performance comparison on COCOA-cls test set}, ResNet-50 as backbone. $\dagger$ indicates our reproduced results.}
    \vspace{-0.5em}
    \centering
    \setlength{\tabcolsep}{4pt}
    \renewcommand{\arraystretch}{.9}
    \resizebox{0.9\textwidth}{!}{
    \begin{tabular}{l|c|c|cccc}
    \toprule
         \multirow{2}{*}{\textbf{Methods}} & \multirow{2}{*}{\textbf{Venue}} & \multirow{2}{*}{\textbf{\shortstack{Shape \\ Prior}}} &  \multirow{2}{*}{$AP\uparrow$} & \multirow{2}{*}{$AP_{50}\uparrow$} & \multirow{2}{*}{$AP_{75}\uparrow$} & \multirow{2}{*}{$AR\uparrow$} \\
  ~ & ~ & ~  &  &  &  &  \\ \midrule
                Mask R-CNN\cite{he2017mask} & ICCV17 & \xmark  & $33.67$ & $56.50$ & $35.78$ & $34.18$ \\ 
        ORCNN\cite{follmann2019learning} & WACV19 & \xmark &  $28.03$ & $53.68$ & $25.36$ & $29.83$ \\ 
        ASN\cite{qi2019amodal}  $\dagger$ & CVPR19 & \xmark & $35.33$ & $58.82$ & $37.10$ & $35.50$ \\ 
        BCNet\cite{ke2021deep} $\dagger$ & CVPR21 & \xmark &  $35.14$ & $\underline{58.84}$ & $36.65$ & $35.80$ \\ 
        AISFormer\cite{tran2022aisformer} & BMVC22 & \xmark  & $\underline{35.77}$ & $57.95$ & $38.23$ & $36.71$ \\        
        VRSP-Net\cite{xiao2021amodal}  & AAAI21 & \checkmark  & $35.41$ & $56.03$ & ${38.67}$ & $\underline{37.11}$ \\ 
        C2F-Seg\cite{gao2023coarse}$\dagger$  & ICCV23 & \checkmark  & $35.72$ & $58.80$ & $\underline{38.73}$ & $\underline{37.11}$ \\ \midrule 
        \textbf{AISDiff (Ours)} & - & \checkmark & \textbf{35.93} & \textbf{58.86} & \textbf{38.63} & \textbf{37.14} \\ \bottomrule
    \end{tabular}
    }
    \label{tab:cocoa-cls}
\end{table}

\subsubsection{Qualitative Results}

\begin{figure}[!h]
    \centering
    \includegraphics[width=\linewidth]{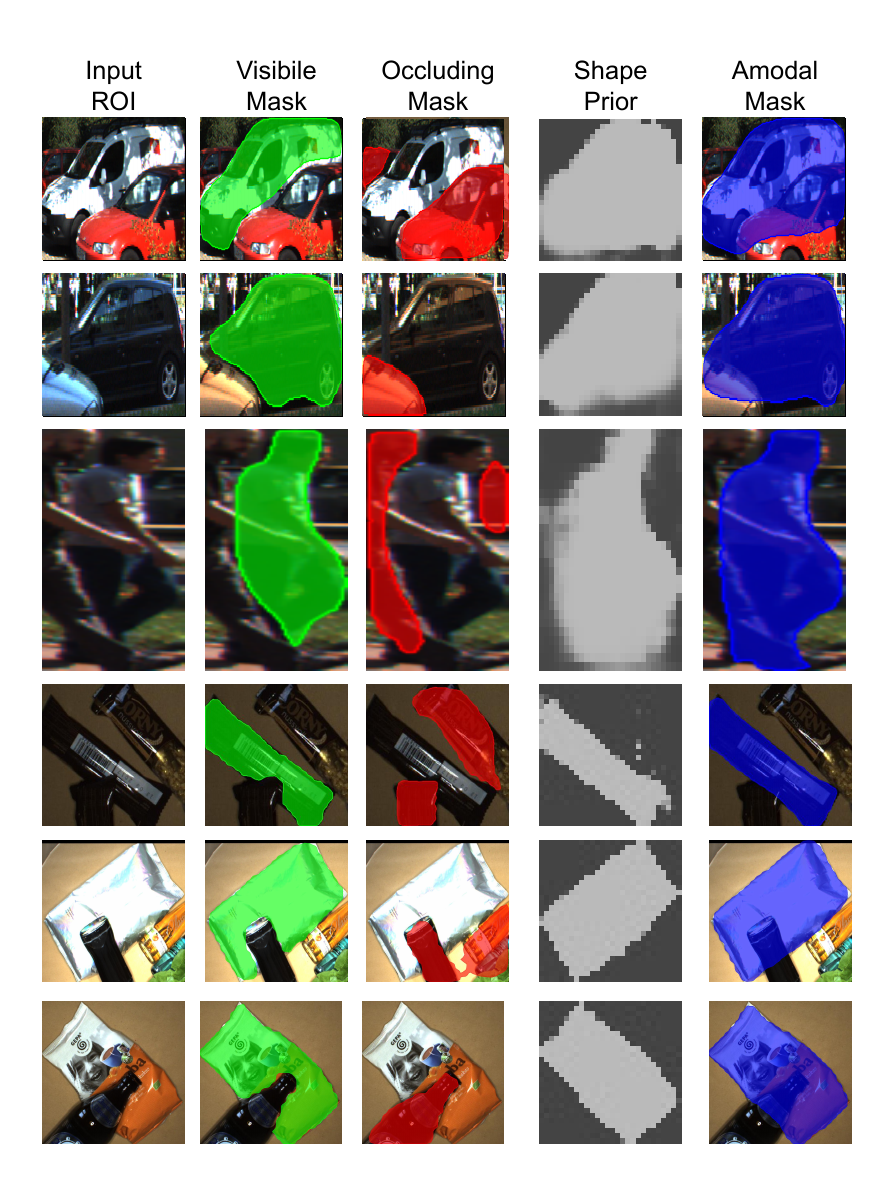}
    \caption{Qualitative results of AISDiff. Left to right: Input RoI, Visible masks, Occluding masks, Amodal masks. Best viewed in color.}
    \label{fig:quali_results}
\end{figure}

\begin{figure}[!t]
    \centering
    \includegraphics[width=.9\textwidth]{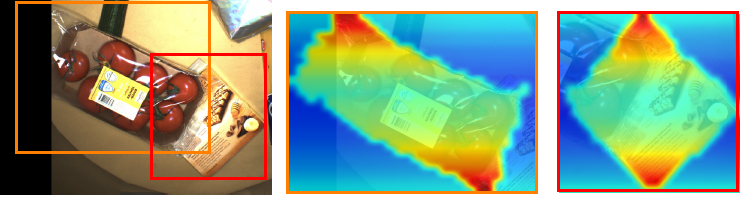}
    \vspace{-0.3cm}
    \caption{Spatial attention map of the Shape Prior Amodal Predictor on the each RoI. Best viewed in color.} 
    \label{fig:attention_viz}
\end{figure}

\begin{figure}[!t]
    \centering
    \includegraphics[width=\textwidth]{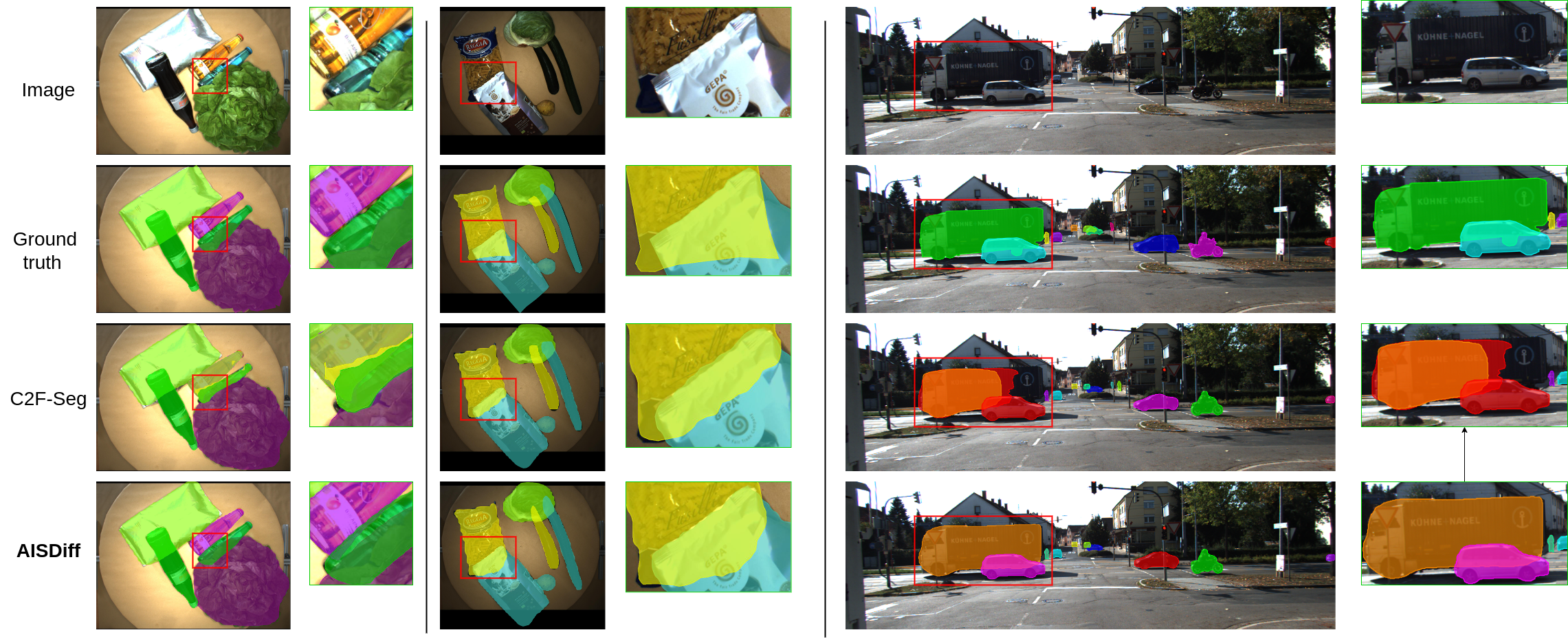}
    \vspace{-0.3cm}
    \caption{Qualitative comparison between AISDiff and the SOTA method C2F-Seg\cite{gao2023coarse}. Examples are from D2SA (left) and KINS (right) test set.} 
    \label{fig:quali_compare}
\end{figure}

\cref{fig:quali_results} illustrates the qualitative output of AISDiff. The results are arranged from left to right, encompassing: input ROIs, Visible Masks, Occluding Masks, Shape Prior, and Amodal Masks.
\cref{fig:attention_viz} visualizes the spatial attention map of the Shape Prior Amodal Predictor on ROIs of the image. 
The attention maps are well-constrained to the object shape.
Moreover, we can see that the decoder typically attends to the visible parts of objects that are similar to the occluded regions when predicting the amodal mask.
\cref{fig:quali_compare} shows
qualitative comparison between AISDiff and the existing SOTA method C2F-Seg\cite{gao2023coarse}.
Example are sampled from D2SA and KINS test sets. As can be seen, AISDiff accurately extracts the amodal mask of the occluded object (i.e. the bag of pasta) (left) and efficiently handles the car and the truck (right).

\subsection{Ablation studies}
\subsubsection{Effect of DiffSP diffusion models}
Table \ref{abla:diffusion_models} compares the performance of using GLIDE\cite{nichol2021glide} and Stable Diffusion\cite{rombach2022high} models within the DiffSP framework on the KINS and D2SA datasets. For the KINS dataset, GLIDE achieves an AP of $35.16$ and an AR of $21.71$, while Stable Diffusion shows improved performance with an AP of $36.36$ and an AR of $22.02$. Similarly, on the D2SA dataset, GLIDE records an AP of $70.18$ and an AR of $69.22$, whereas Stable Diffusion further excels with an AP of $71.01$ and an AR of $69.29$. These results indicate that Stable Diffusion consistently outperforms GLIDE, offering better precision and recall in the DiffSP framework.
\begin{table}[!h]
    \setlength{\tabcolsep}{4pt}
    \centering
    \caption{Effect of diffusion models in DiffSP}
    \begin{tabular}{l|cccc|cccc}
    \toprule
        \multirow{2}{*}{\textbf{Diffusion Model}} & \multicolumn{4}{c}{\textbf{KINS}} & \multicolumn{4}{c}{\textbf{D2SA}} \\ \cmidrule{2-9}
         & $AP\uparrow$ & $AP_{50}\uparrow$ & $AP_{75}\uparrow$ & $AR\uparrow$ & $AP\uparrow$ & $AP_{50}\uparrow$ & $AP_{75}\uparrow$ & $AR\uparrow$ \\ \midrule 
        GLIDE\cite{nichol2021glide} & 35.16 & 57.97 & 37.11 & 21.71 & 70.18 & 85.11 & 74.96 & 69.22 \\ 
        Stable Diffusion\cite{rombach2022high} & 36.36 & 58.84 & 37.24 & 22.02 & 71.01 & 85.12 & 76.23 & 69.29 \\ \bottomrule
    \end{tabular}
    \label{abla:diffusion_models}
\end{table}

\subsubsection{Effect of denoising time steps} Table \ref{abla:denoising_timestep} presents the impact of varying diffusion timesteps (T = 10, 50, 100) on the performance of the DiffSP framework, evaluated using the Average Precision (AP) and Average Recall (AR) metrics on the KINS and D2SA datasets. For the KINS dataset, the AP remains relatively stable across different timesteps, with values of $36.36$, $36.33$, and $36.43$ for $T=10$, $T=50$, and $T=100$, respectively. Similarly, the AR values are consistently around $37.24$ for all timesteps. In the D2SA dataset, the AP values are $71.01$, $70.99$, and $71.11$ for $T=10$, $T=50$, and $T=100$, respectively, while the AR values remain steady at approximately 69.29 across all timesteps. Overall, the results indicate that varying the number of diffusion timesteps has minimal impact on both AP and AR metrics for both datasets, suggesting that the DiffSP framework performs robustly regardless of the diffusion timestep settings. Thus, we opt for $T=10$ for efficiency.
\begin{table}[!h]
    \centering
    \setlength{\tabcolsep}{8pt}
    \caption{Effect of difusion timesteps}
    \begin{tabular}{l|cccccc}
    \hline
        \multirow{2}{*}{\textbf{Dataset}} & \multicolumn{2}{c}{$T=10$} & \multicolumn{2}{c}{$T=50$} & \multicolumn{2}{c}{$T=100$} \\ \cmidrule{2-7}
        ~ & $AP$ & $AR$ & $AP$ & $AR$ & $AP$ & $AR$ \\ \midrule 
        \textbf{KINS} & 36.36 & 37.24 & 36.33 & 37.23 & 36.43 & 37.24 \\ 
        \textbf{D2SA} & 71.01 & 69.29 & 70.99 & 69.27 & 71.11 & 69.29 \\ \hline
    \end{tabular}
    \label{abla:denoising_timestep}
\end{table}

\subsubsection{Effect of object category and occluding mask}
We conducted an ablation study on the inputs of DiffSP, which utilizes three inputs: visible pixels of the object of interest, the occluding mask, and the object category. 
The first input is essential for reconstructing the missing parts and cannot be omitted. Therefore, our study focuses on the other two inputs: the object category and the occluding mask, as shown in Table \ref{abla:diff_sp_input}. 
The results demonstrate that including either the object category or the occluding mask improves performance, with the best results achieved using both (the default DiffSP configuration).

\begin{table}[!h]
    \vspace{-1em}
    \caption{Effect of object category and occluding mask}
    \centering
    \setlength{\tabcolsep}{8pt}
    \label{tab:model_efficiency}
    \begin{tabular}{c|c|cc|cc}
    \toprule
        \multirow{2}{*}{\shortstack{\textbf{Object}\\ \textbf{Category}}} & \multirow{2}{*}{\shortstack{\textbf{Occluding} \\\textbf{Mask}}} & \multicolumn{2}{c}{\textbf{KINS}}  & \multicolumn{2}{c}{\textbf{D2SA}} \\ \cline{3-6}
        ~ & ~ & \textbf{AP}$\uparrow$ & \textbf{AR}$\uparrow$ & \textbf{AP}$\uparrow$ & \textbf{AR}$\uparrow$ \\ \midrule
        \xmark & \xmark & 31.12 & 20.10 & 64.87 & 67.09 \\ 
        \checkmark & \xmark & 35.23 & 21.83 & 67.92 & 68.13 \\ 
        \xmark & \checkmark & 35.81 & 21.87 & 69.13 & 69.21 \\ 
        \checkmark & \checkmark & 36.36 & 22.02 & 71.01 & 69.29\\ \bottomrule
    \end{tabular}
    \label{abla:diff_sp_input}
    \vspace{-2em}
\end{table}

\section{Conclusion} 
In conclusion, we propose AISDiff, an AIS mask head with a Diffusion Shape Prior Estimation module. This module, termed DiffSP, leverages pre-trained conditioned diffusion models on extensive datasets to extract nuanced visual features for deriving the shape prior of the object. Furthermore, we present the Shape Prior Amodal Predictor, which utilizes attention-based feature maps from the shape prior to enhance amodal segmentation. Through extensive experimentation across diverse AIS benchmarks, we affirm the efficacy of AISDiff.

\begin{credits}
\subsubsection{\ackname} This work is sponsored by the National Science Foundation (NSF) under Award No OIA-1946391.
\end{credits}

\bibliographystyle{splncs04}
\bibliography{main}
\end{document}